\documentclass[letterpaper, 10 pt, conference]{ieeeconf}  % Comment this line out if you need a4paper

\IEEEoverridecommandlockouts                              % This command is only needed if 
\usepackage{xcolor}
\usepackage{amsmath}
\usepackage{amssymb}
\usepackage{bm}
\usepackage{graphicx}
\usepackage{booktabs}
\usepackage{multirow}
\usepackage{hyperref}
\usepackage{subcaption}
\title{\LARGE \bf
Autonomous Excavation of Challenging Terrain using Oscillatory Primitives and Adaptive Impedance Control}

\author{Noah Franceschini, Pranay Thangeda, Melkior Ornik, Kris Hauser
        \\University of Illinois Urbana-Champaign%
        \\\tt{\{nef3,pranayt2,mornik,kkhauser\}@illinois.edu}%
}

\begin{document}

\maketitle
\thispagestyle{empty}
\pagestyle{empty}

%%%%%%%%%%%%%%%%%%%%%%%%%%%%%%%%%%%%%%%%%%%%%%%%%%%%%%%%%%%%%%%%%%%%%%%%%%%%%%%%
\begin{abstract}
This paper addresses the challenge of autonomous excavation of challenging terrains, in particular those that are prone to jamming and inter-particle adhesion when tackled by a standard penetrate-drag-scoop motion pattern. Inspired by human excavation strategies, our approach incorporates oscillatory rotation elements -- including swivel, twist, and dive motions -- to break up compacted, tangled grains and reduce jamming. We also present an adaptive impedance control method, the Reactive Attractor Impedance Controller (RAIC), that adapts a motion trajectory to unexpected forces during loading in a manner that tracks a trajectory closely when loads are low, but avoids excessive loads when significant resistance is met. Our method is evaluated on four terrains using a robotic arm, demonstrating improved excavation performance across multiple metrics, including volume scooped, protective stop rate, and trajectory completion percentage. 
\end{abstract}

%%%%%%%%%%%%%%%%%%%%%%%%%%%%%%%%%%%%%%%%%%%%%%%%%%%%%%%%%%%%%%%%%%%%%%%%%%%%%%%%
\section{INTRODUCTION}

Autonomous excavation --- which refers to the navigation, loading, and planning of removing or depositing materials without the assistance of a human driver --- is a  promising frontier of robotics \cite{melenbrink2020onsite}. Excavation is an economically important activity for constructing roadways, foundations for buildings, and earth structures in forward military bases, as well as mining and waste handling operations, and automated excavation has been studied for decades since the pioneering work of Singh \cite{singh1995synthesis} in relation to planning \cite{KIM2020103108}, perception \cite{yeom2019surround}, and control \cite{4472850}. This paper focuses on excavation trajectory planning and execution. Whereas standard excavation motions perform well in some terrains, particularly in homogeneous, fine-grained terrains and non-compacted soils, they may fail to yield a sufficient volume or excavation profile in more complex terrains \cite{DADHICH2016212}. In particular, tool-material interactions experienced during an excavation loading cycle may be unpredictable due to jamming phenomena in rocky materials and variability in terrain composition, such as large embedded rocks and tree roots hidden beneath a soil surface~\cite{4472850,marshall2001towards,zhang2023novel}. Moreover, unusual inter-particle interactions like adhesion and snagging occur when particle shapes are highly irregular, such as those observed in organic terrains like mulch, and can significantly reduce excavated volumes. Although human operators observe terrain response and adjust their excavation strategy appropriately for a given terrain, prior work has not adequately addressed these issues in the automated case.

\begin{figure}[tbp]
\centering
\includegraphics[width=0.49\linewidth, ]{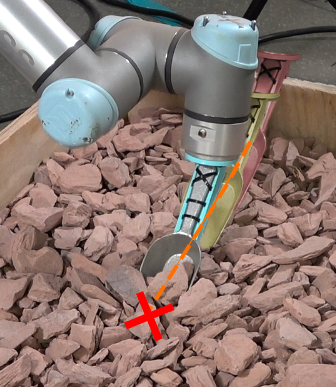} 
\includegraphics[width=0.49\linewidth, ]{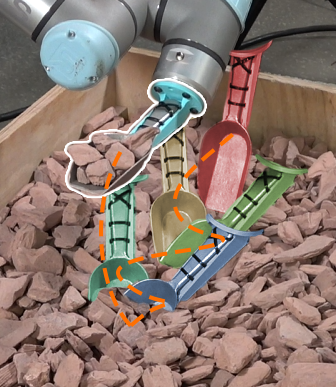} 
\caption{Strobe-effect visualization of a robot excavating large-grained slate. Motion of the scoop is illustrated via tinting progressing from red to blue hue. Left: standard Penetrate-Drag-Scoop (PDS) motion with impedance control fails due to jamming forces triggering a protective stop (red X). Right: using our ``swivel'' primitive and Reactive Attractor Impedance Controller (RAIC), the robot breaks through jammed particles and excavates a large volume (outlined in white). Trajectory of the scoop is illustrated as an orange curve.}
\label{fig:overview}
\end{figure}

The majority of prior autonomous excavation work uses a predefined trajectory class: penetrate-drag-scoop (or penetrate-drag-curl), in which an excavator bucket first penetrates the terrain with its leading edge, the bucket is dragged through the terrain, and the bucket is rotated to lift the material away from the terrain \cite{9484815}. The penetration angle, depth, and drag length can be adjusted for a given terrain.  Inspired by observations of human adjustments of excavation strategies in such terrains, we introduce a class of variations to the basic primitive that uses oscillations to break up grains to reduce the rate of jamming and encourage disentangling between irregular grains.  We empirically examine the effect of sinusoidal rotation perturbations along different axes, frequencies, and amplitudes as the scoop penetrates the terrain.

Moreover, although impedance control \cite{4788393} is a common strategy to introduce compliance in excavation \cite{ha2000impedance, 7962150}, standard methods of impedance control are still prone to jamming and exceeding safe force limits in challenging terrain. We introduce a novel adaptive method for trajectory tracking control, Reactive Attractor Impedance Controller (RAIC), that alleviates the jamming problem found in the loading cycle by adapting the progress of the impedance control attractor along the tracked path according to the experienced resistance. At low resistance the attractor moves at full speed, and slows if resistance is high.  Our experiments evaluate our approach in a variety of terrains, demonstrating that the combination of oscillatory primitives and adaptive impedance significantly increase excavated volumes and reduce protective stop rates in challenging terrain.  Moreover, the RAIC controller can even adapt to hidden immovable obstacles embedded in the terrain.

\section{Related Work}

\subsection{Excavation}
Automating excavation is a long-standing and ongoing effort towards creating more efficient construction worksites \cite{ERALIEV2022104428,Bradley1998}. Previous studies have shown that autonomous excavation can increase fuel efficiency \cite{Yao2023BucketLT}, decrease load time \cite{Dobson2016, 8972589}, and replace human workers in dangerous environments such as underground-mine excavation \cite{app11188718}.  Planning, perception, control, and learning are the key ingredients of autonomous excavation systems~\cite{singh1995synthesis}.  Although much interesting work addresses high-level perception and decision making for orchestrating navigation and excavation locations~\cite{KIM2020103108,yeom2019surround}, we focus in this paper on lower-level problems of trajectory planning and control once an excavation location, depth, and drag length have been chosen. 

\subsection{Excavator Trajectory Planning}
Yao et al. \cite{Yao2023BucketLT} use a trajectory optimization method to fill a Load-Haul-Dump machine bucket to a desired fill factor while minimizing fuel usage and constraining the optimization to safe torque limits. While this trajectory optimization method proved to create more fuel efficient trajectories, the work assumes the soil is homogeneous and requires prior knowledge on the composition of the material. Fragmented rock has been identified as a challenge for standard excavation approaches~\cite{lu2022excavation,zhu2022excavation,4472850}.  Vision-based learning techniques~\cite{lu2022excavation,zhu2022excavation} have been applied to identify areas between rocks in which the bucket edge should begin excavating. However, these learning approaches focused on finding location and shape parameters for the standard penetrate-drag-scoop primitive rather than maneuvering the scoop.

%\subsection{Manipulation of Granular Media}

%The manipulation of granular media \cite{Tuomainen_2022} has a wide range of real-world applications such as terrain shaping \cite{pmlr-v78-schenck17a}, scooping \cite{8610813}, food packaging \cite{DBLP:journals/corr/abs-2105-12946}, excavation \cite{Bradley1998}, etc. In Schenck et al.~\cite{pmlr-v78-schenck17a}, they use two convolutional neural networks to predict the outcome of scooping and dumping actions. They then use these predictions to create a sequence of actions, which are defined as scoop poses, and linearly interpolate between them to shape the granular media to a desired height map. The focus of this study is more related to the terrain shaping itself and where to scoop rather than more effectively scooping the material that the robotic arm is shaping. Training these models also required tens of thousands of sample scoops which can be costly in time. In \cite{DBLP:journals/corr/abs-2105-12946}, they use a neural network to predict the mass of granular food particles (such as beans, rice, oatmeal, etc.) that a grasping action would yield at a specific position. Again, this paper focuses more on where to take a scooping/grasping action rather than a study of the way that the granular material can more effectively be scooped. Both of these studies are also limited to smaller grain sizes comparable to pebbles or smaller and do not perform any studies on performing these actions on larger materials.

\subsection{Excavation Control}
Open-loop excavation control can be dangerous during jamming or when encountering unexpected obstacles, as high forces can cause damage to actuators and or tip an excavator over. Impedance control~\cite{ha2000impedance, 4788393,FengAdaptive} and admittance control~\cite{4472850,fernando2019iterative} have been proposed to modify the controls and comply to unexpected terrain resistance.  Here, the robot tracks the end effector toward an attractor pose, but also responds to measured forces using a mass-damper simulation~\cite{hogan1984impedance}.  Controller parameters can be learned iteratively by observing excavated fill weight and adjusting parameters to help obtain a desired weight for novel terrain materials~\cite{fernando2019iterative}. Although standard impedance / admittance control achieves compliance to unexpected forces, it can still encounter large forces when the distance between the current and target pose grows large due to substantial obstructions. 
Azulay et. al. recently proposed an end-to-end deep reinforcement learning model that directly controls the velocity, bucket tilt and lift of an LHD machine~\cite{9344588}. Egli et al.~\cite{ETH} also propose an RL model that directly drives joint velocities of a wheeled loader. The RL policy is rewarded for quick and filling scoops and negatively rewarded for erratic movements and self collisions. Although this study provides promising results for RL controlled excavation, the work is mostly limited to soft, homogeneous soils.  Our proposed controller is an adaptation of impedance control tailored for severe cases of jamming, and it requires far less tuning than RL-based methods.

%%%%%%%%%%%%%%%%%%%%%%%%%%%%%%%%%%%%%%%%%%%%%%%%%%%%%%%%%%%%%%%%%%%%%%%%%%%%%%%
% \section{Problem Statement}
% Given an arbitrary terrain $T$ composed of granular or irregular materials, we aim to maximize the volume of material V scooped in a single excavation cycle while minimizing the maximum force $F_{max}$ experienced by the robot's end-effector. The excavation cycle is defined by a start position $p_start$, a penetration depth $d$, and a drag length $l$. Formally, we seek to solve:
% \begin{equation*}
%     \max_{s \in S} V(s, T)
% \end{equation*}
% \begin{equation*}
%     \text{subject to } F_{max}(s, T) \leq F_{threshold}
% \end{equation*}
% where $S$ is the set of all possible scooping strategies, $V(s, T)$ is the volume scooped using strategy $s$ on terrain $T$, and $F_{max}(s, T)$ is the maximum force experienced during the execution of strategy $s$ on terrain $T$. $F_{threshold}$ is a predefined force limit to ensure safe operation.

% The challenge lies in developing a scooping strategy s that is robust and adaptable to various terrain compositions, including those prone to jamming or containing irregular materials, without prior knowledge of the specific terrain properties.

%%%%%%%%%%%%%%%%%%%%%%%%%%%%%%%%%%%%%%%%%%%%%%%%%%%%%%%%%%%%%%%%%%%%%%%%%%%%%%%%
\section{Methodology}
Motivated by observations of human strategies for excavating challenging terrains, we present parameterized scooping primitives that enhance the standard penetrate-drag-scoop (PDS) trajectory, and the  Reactive Attractor Impedance Controller (RAIC) which enhances standard impedance control.

\subsection{Parameterized Scooping Primitive}
\begin{figure}[tbp]
    \centering
    \includegraphics[trim=0.5cm 6cm 10.5cm 2.1cm,clip,width=0.9\linewidth]{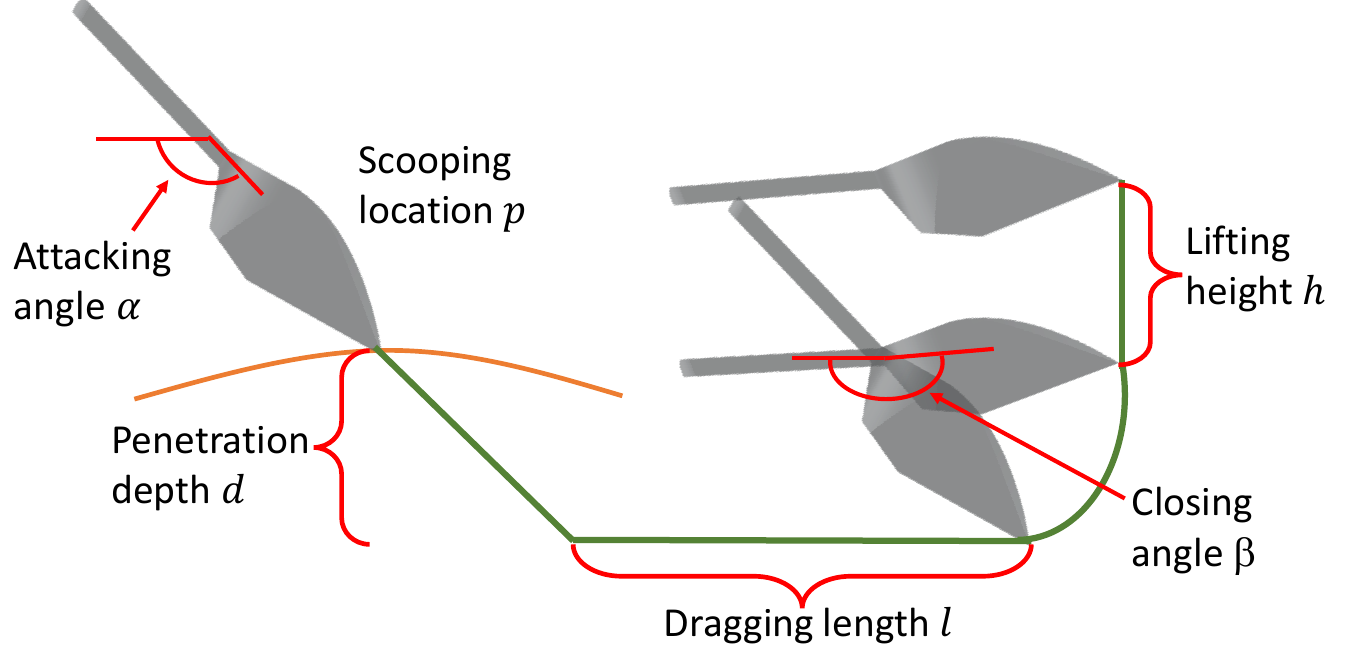}
    \caption{Illustration of the PDS trajectory.}
    \label{fig:pds_trajectory}
\end{figure}

The penetrate-drag-scoop (PDS) trajectory, illustrated in Fig.~\ref{fig:pds_trajectory}, is a three-phase loading cycle widely employed by earth-moving vehicles such as backhoes, wheeled loaders, and excavators. This trajectory begins at a point \textit{p} and moves in a plane with $x$ direction aligned to a heading $\theta_\text{base}$ and $y$ direction vertical. The scoop will also pitch in a prescribed manner about the axis perpendicular to this plane. The cycle begins with a penetration phase, where the end-effector enters the target material at an attack angle $\alpha$, reaching a depth \textit{d}. Subsequently, in the dragging phase, the scoop maintains its initial attack angle while being dragged linearly for a distance \textit{l}. The final scooping phase involves rotating the scoop to a closing angle $\beta$, followed by a vertical ascent to height \textit{h}, thus extracting the scooped material. The trajectory can be described as:
\begin{equation}
    \mathbf{T}_\text{PDS}(t) = \begin{cases}
        \mathbf{T}_\text{penetrate}(t), & 0 \leq t < t_1 \\
        \mathbf{T}_\text{drag}(t), & t_1 \leq t < t_2 \\
        \mathbf{T}_\text{scoop}(t), & t_2 \leq t \leq t_3
    \end{cases}
\end{equation}
where $\mathbf{T}(t) \in SE(3)$ represents the scoop tip pose in the world frame at time $t$, and $t_1$, $t_2$, and $t_3$ are the transition times between phases.

Although PDS is effective at scooping fine-grained granular media, it suffers in terrains where penetration is difficult due to jamming and in terrains with high inter-particle friction forces. To help alleviate these inter-particle forces, we extend the standard PDS trajectory with three novel scooping primitives inspired by motions humans use when scooping materials: the swivel, the twist, and the dive. These primitives introduce variable frequency and amplitude perturbations designed to improve performance in terrains prone to jamming or containing irregular materials.

\begin{figure}[tbp]
    \centering
    \includegraphics[width=0.98\linewidth]{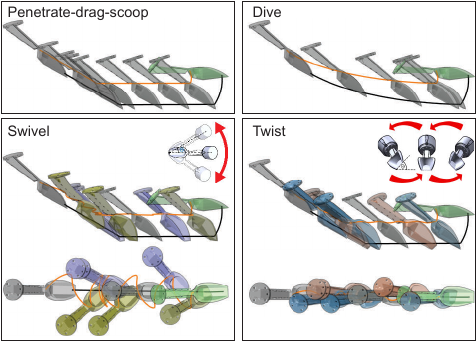}
    \caption{Illustration of our proposed primitives.  Bottom row shows side and t
    op-down views of oscillatory trajectories.  Scoop tint indicates magnitude of roll (cyan - orange), yaw (blue - yellow), and pitch (green).}
    \label{fig:all_primitives}
\end{figure}

\subsubsection{Swivel Primitive}
Swiveling introduces an oscillatory rotation about the scoop's tip, aiding in breaking inter-particle friction and disrupting jammed particles when the scoop tip encounters resistance. It is defined  by amplitude $A_\text{swivel}$ and frequency $f_\text{swivel}$ with yaw defined by:
\begin{equation}
\theta_\text{swivel}(t) = A_\text{swivel} \sin(2\pi f_\text{swivel} t).
\end{equation}
This yaw value is added to the base yaw $\theta_\text{base}$ from the beginning of the penetration phase to the end of the dragging phase. During the scoop phase we return the yaw to $\theta_\text{base}$.
 
\subsubsection{Twist Primitive}
Twisting introduces an oscillatory rotation about the scoop's longitudinal axis, and helps disentangle particles outside of the scoop from dragging out particles inside of the scoop, which is particularly effective in materials with high inter-particle adhesion. The roll angle is parameterized by amplitude $A_\text{twist}$ and frequency $f_\text{twist}$ as:
\begin{equation}
\phi_\text{twist}(t) = A_\text{twist} \sin(2\pi f_\text{twist} t)
\end{equation}
Like the swivel, twisting is only active from the beginning of the penetration phase through the end of the drag phase. During the scoop phase the roll is returned back to 0.

\subsubsection{Dive Primitive}
Diving modifies the translation and pitch of the penetration and drag phases of the PDS trajectory, creating a smooth, continuous curve. It is parameterized by a {\em dive curve} factor $s \in [0, 1]$, where $s = 0$ corresponds to the original PDS trajectory and $s = 1$ generates a fully smooth curve. The dive trajectory modifies the drag phase such that the drag phase becomes replaces by this smoothed trajectory as $s$ increases to one.  We define $\mathbf{p}_\text{smoothed}(u)$ as a quadratic B\'ezier curve using the penetration point, the original starting pose of dragging phase $p_{drag start}$ and drag length $s \cdot l$ from $p_{drag start}$ as control points. Given the parameter $s$, the trajectory of the end effector is given by:
\begin{equation}
    \mathbf{p}_\text{dive}(u) = (1-s)\mathbf{p}_\text{drag}(u) + s\mathbf{p}_\text{smoothed}(u)
    \label{dive_eq}
\end{equation}
where $\mathbf{p}_{\text{drag}}$ is the translation defined by the original PDS $\mathbf{T}_\text{drag}$ trajectory, and $u$ is the normalized trajectory parameter. To smooth out and replace the drag phase as $s$ increases, we linearly shrink the drag phase by $(1-s)*l$ such that at $s = 1$ the drag phase is fully replaced by $p_{\text{smoothed}}$ with $p_{smoothed}$ using control points $p$, $p_{drag start}$, and drag length $l$ from $p_{drag start}$. 
In addition to smoothing translation, the dive curve modifies the pitch of the scoop to keep the scoop edge traveling tangent to the curve. The pitch at time \textit{t} is defined as follows:
\begin{equation}\label{pitch_eq}
    \text{pitch}(t) = tan^{-1}(p^\prime_{z}/ \sqrt{\Delta p_{x}^{\prime2} + p_{y}^{\prime 2}})
\end{equation}
This modification facilitates more effective media penetration and encourages particle ingress into the scoop during the loading cycle. 

We found that this primitive is most useful when paired with the swivel or twist motions as keeping the scoop tangent to its motion allows for the oscillatory movements to more easily cut through the materials.

These primitives can be applied individually or in combination, allowing for adaptive scooping strategies tailored to specific terrain characteristics.  
Given a set of desired parameters $(A_{\text{swivel}},f_{\text{swivel}},A_{\text{twist}},f_{\text{twist}},s)$, the three primitives are combined to provide the pose trajectory for the scoop as follows. First, we convert our roll, pitch and yaw into a combined rotation matrix, defined as:
\[
R_{\text{prim}}(t) = R_{z}(\theta_\text{base}+\theta_{\text{swivel}}(t)) \cdot R_{y}(\text{pitch}(t)) \cdot R_{x}(\phi_\text{twist}(t))
\]
where $R_x$, $R_y$, and $R_z$ are rotation matrices about the elementary axes.
The translation is defined by $\mathbf{p}_{\text{dive}}$ \eqref{dive_eq}, which is identical to the PDS translation trajectory if $s=0$. In the scoop phase, the pitch linearly interpolates from pitch($t_2$) to closing angle $\beta$ to finish the scooping trajectory, and the swivel and twist are interpolated to 0 to keep the scoop level upon termination.

\subsection{Reactive Attractor Impedance Controller}
Our RAIC controller is a modification of a standard impedance controller that avoids applying excessive forces while following a target trajectory. RAIC behaves exactly like an impedance controller does in zero-to-little resistance materials, but limits how far the attractor moves away from the end effector when it experiences a lack of progression in medium-to-high resistance materials. 

Standard impedance control attracts the current end-effector pose $\mathbf{T}$ to the planned pose $\mathbf{T}^{\text{plan}}$ via a simulating a spring-damper system.  Instead, RAIC evolves an attractor pose $\mathbf{T}^{\text{att}}$ to move along the planned trajectory when loads are low, but allows deviations when they are high, which avoids approaching the robot's force / torque limits.  Moreover, our control scheme allows the attractor point to become ``unstuck'' if the plan pulls it in a direction that may relieve currently accumulated load.

More specifically, we define $\mathbf{T}_t \in SE(3)$ as the current end-effector pose at time $t$, and $\mathbf{T}^{\text{plan}}_{t}$ the planned pose, and attractor pose as $\mathbf{T}_0^\text{att}$.  Let $\mathbf{f} = [f_x, f_y, f_z, \tau_x, \tau_y, \tau_z]^\top \in \mathbb{R}^6$ be the vector of measured forces and torques from a force/torque sensor.  We also define two elementary operators. The first computes the difference vector between poses as 
\begin{equation}
    d(\mathbf{T}^\prime,\mathbf{T}) = [\omega (\mathbf{R}^\prime)^{-1} \mathbf{R}, \mathbf{p}^\prime - \mathbf{p}] \in \mathbb{R}^6
\end{equation}
where the notation $\mathbf{R} \in SO(3)$ and $\mathbf{p} \in \mathbb{R}^3$ split a pose into its rotation and position components, and $\omega(R)$ gives the rotational displacements of $R$ about the $x$, $y$, and $z$ axes.  Conversely, the integration operator $int(\mathbf{T},\psi)$ to apply a difference vector $\psi$ to a pose. This operator acts as an inverse of $d$, such that $\mathbf{T}^\prime = int(\mathbf{T}, d(\mathbf{T}^\prime,\mathbf{T}))$.

The attractor begins at $\mathbf{T}_0^\text{att} = \mathbf{T}_0$ and evolves along $\mathbf{T}^\text{plan}$ according to a feedforward and feedback component.  Specifically, we compute the difference in the planned pose 
\begin{equation}
\mathbf{d} = d(\mathbf{T}^\text{plan}_{t+1},\mathbf{T}^\text{plan}_{t})
\end{equation}
and the difference from the planned pose to the attractor position
\begin{equation}
\mathbf{e} = d(\mathbf{T}^\text{plan}_{t},\mathbf{T}^\text{att}_{t}).
\end{equation}
The attractor update is given by a damped proportional control with feedforward and feedback terms:
\begin{equation}
    \mathbf{T}^\text{att}_{t+1} = int(\mathbf{T}^\text{att}_{t}, D(\mathbf{f},\mathbf{e})\mathbf{d}^\prime + F(\mathbf{f},\mathbf{e})\mathbf{e}).
\end{equation}
Here, $\mathbf{d}^\prime$ is a small modification to $\mathbf{d}$ that discards feedforward directions that close the gap between the plan and the attractor:
\begin{equation*}
    d_i = d_i \cdot \min(1+sign(d_i \cdot e_i),1).
\end{equation*}
The gains $D$ and $F$ are defined such that $D=I$ and $E=I$ when resistive forces are low, and they are close to 0 when resistive forces are high and the planned differences are pointing opposite to the felt forces.

Specifically, we define a damping function $\phi_i: \mathbb{R} \rightarrow [0,1]$ for each axis $i$:
\begin{equation}\label{Damping function}
    \phi_i(f_i) = \text{clip}(1 - s_i |f_i - c_i|, 0, 1)
\end{equation}
where $s_i > 0$ is a scaling factor and $c_i$ is a cutoff value for axis $i$. $\phi_i$ creates a piecewise linear damping effect that begins at 1, starts decreasing linearly at $c_i$, and saturates at 0.

We also define an ``exit'' damping function $\varphi_i \rightarrow [0,1]$ for each axis $i$:
\begin{equation*}\label{exit_func}
    \varphi_i(f_i,e_i) = \max(0, \tanh(f_i \cdot e_i))
\end{equation*}
Where $e_i$ is the corresponding element of $\mathbf{e}$.
$\varphi_i$ is 0 when the direction from the attractor to the plan opposes the direction of sensed forces, and rises to 1 when they are in the same direction.

Overall, the feedforward damping function is
\begin{equation*}
    \lambda_i(f_i, e_i) = \max(\phi_i(f_i),\varphi_i(f_i,e_i))
\end{equation*}
and the damping matrix $\mathbf{D} \in \mathbb{R}^{6 \times 6}$ is
\begin{equation}
    \mathbf{D} = \text{diag}(\lambda_1(f_1,e_1), \ldots, \lambda_6(f_6,e_6)).
    \label{fullDamp}
\end{equation}

For the feedback gain, we use a similar piecewise linear  damping.  Governed by feedback scales $s^{\text{fb}}_i$ and cutoffs $c^{\text{fb}}_i$, we define:
\begin{equation*}
    \phi^{fb}_i(f_i) = \text{clip}(1 - s^{fb}_i |f_i - c^{fb}_i|, 0, 1)
\end{equation*}
and
\begin{equation*}
    \lambda^{fb}_i(f_i, e_i) = \max(\phi^{fb}_i(f_i),\varphi_i(f_i,e_i))
\end{equation*}
giving the feedback damping matrix
\begin{equation}
    \mathbf{F} = \text{diag}(\lambda^{fb}_1(f_1,e_1), \ldots, \lambda^{fb}_6(f_6, e_6))
    \label{fullFbDamp}
\end{equation}

The attractor is used as the target for a standard impedance controller. A spring-damper system is simulated as follows. First, given the difference from the current pose to the attractor,
\begin{equation*}
    \mathbf{e}^{spring} = d(\mathbf{T}^\text{att}_{t}, \mathbf{T}_t),
\end{equation*}
We simulate a spring-damper with $\mathbf{f}$ as an external force:
\begin{equation}
    M \ddot x + B \dot{\mathbf{e}}^{spring} + K \mathbf{e}^{spring} = \mathbf{f}.
    \label{spring-damper}
\end{equation}
Here $M$ is the inertia matrix, which we set to an identity, $B$ is the damping matrix, and $K$ is the spring stiffness. We set $B$ and $K$ to diagonal matrices with isotropic behavior in rotation and in translation. Solving for the acceleration $\ddot x$, we then double integrate $\ddot x$ to find a difference $\Delta x$, which is then added to the end-effector pose:
\begin{equation*}
    \mathbf{T}_{t+1} = int(\mathbf{T}_t,\Delta x).
\end{equation*}

\section{Experiments and Results} 

\subsection{Setup Overview}
Our experimental testbed consists of a Universal Robots UR5e robot arm equipped with a metallic scoop as its end effector. We use the built-in force-torque sensing of UR5e for implementing impedance control as well as RAIC.  As a cobot, the UR5e implements a notion of a {\em P-stop} (protective stop) that detects potential damage and halts motion. We use P-stop rate as a proxy for a high-force event in which a robot or construction machinery could experience irreversible damage. 

An Intel RealSense L515 RGB-D camera mounted above the workspace is used to estimate the volume of excavated material. Excavation occurs in a filled wooden box measuring 0.9\,m x 0.6\,m x 0.2\,m and placed so that the robot is unlikely to reach the end of its workspace.
We selected four terrain materials with a diverse set of physical properties: Pebbles, Gravel, Slate, and Mulch. Pebbles, with particles 3-8 mm in diameter, introduces moderate challenges in terms of inter-particle friction and potential jamming. Mulch present unique difficulties due to their irregular shape and tendency to interlock, testing our system's ability to handle materials with high inter-particle friction and potential for entanglement. Gravel and Slate, approximately 2-5 cm in size, represent the most challenging terrain, prone to severe jamming and requiring substantial force to manipulate. 

% \begin{figure}[h!]
% \centering
% \includegraphics[width=0.8\linewidth]{figures/figure_4_small.png}
% \caption{Our experimental setup.}
% \label{fig:overview}
% \end{figure}

\begin{table}[tbp]
    \centering
    \begin{tabular}{c@{\hskip 1mm}c@{\hskip 1mm}c@{\hskip 1mm}c}
     {\bf Pebbles} & {\bf Gravel} & {\bf Slate} & {\bf Mulch} \\
    \includegraphics[width=0.235\columnwidth]{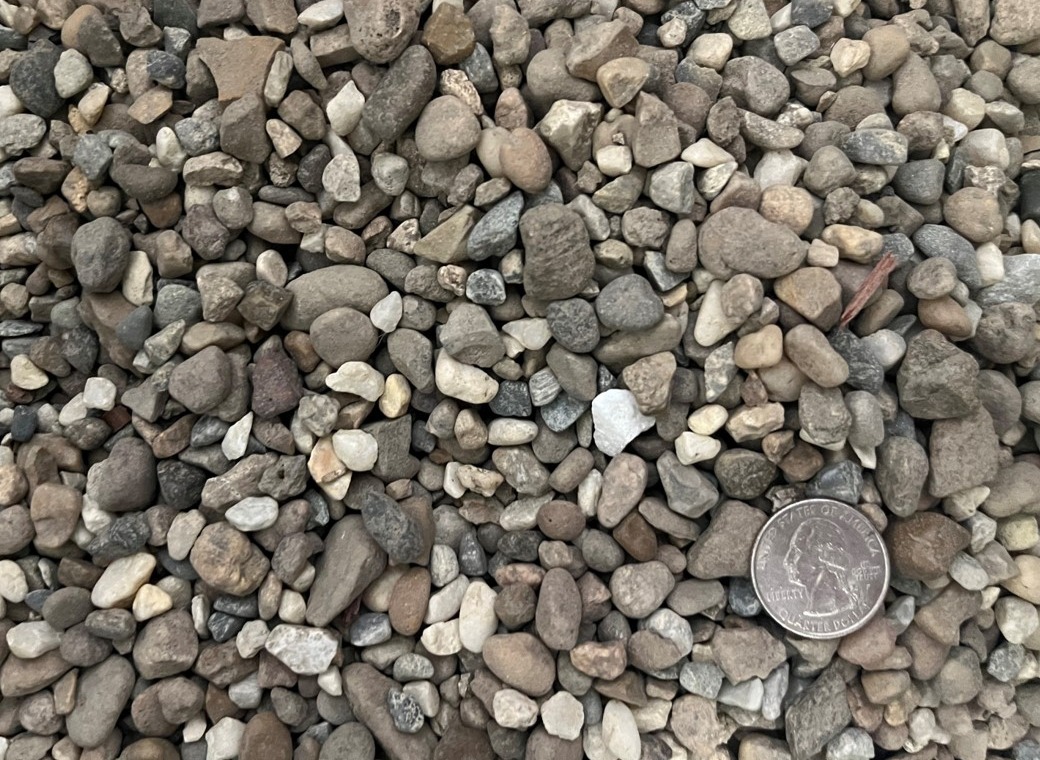}
      & \includegraphics[width=0.235\columnwidth]{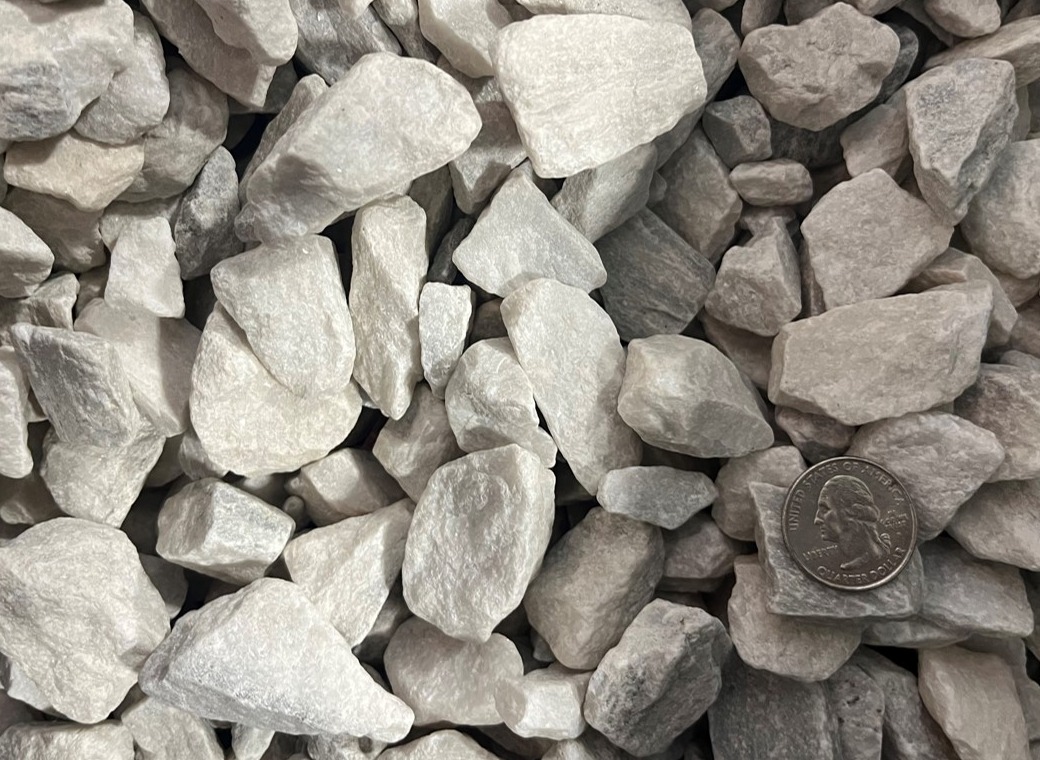}
      & \includegraphics[width=0.235\columnwidth]{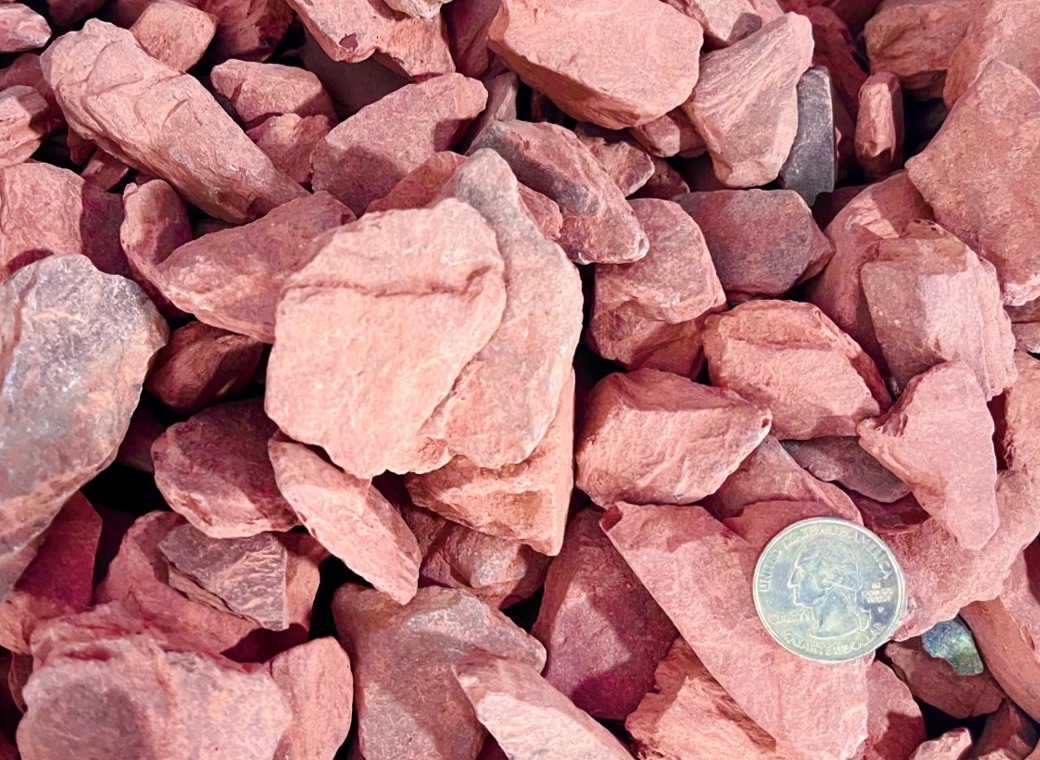}
      & \includegraphics[width=0.235\columnwidth]{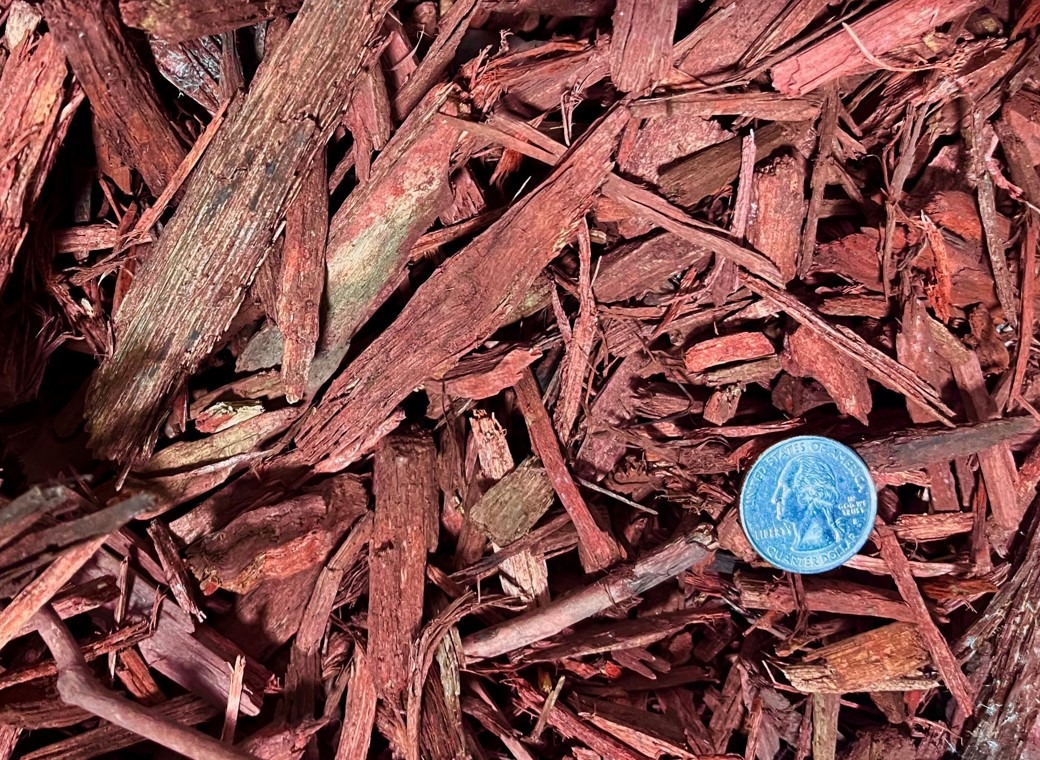}
     \end{tabular}%
    \caption{Materials in test terrains, with U.S. quarter coin shown for scale. }
    \label{tbl:materials}
\end{table}

\subsection{Controller Parameter Tuning}

Our methodology for choosing the parameters of our impedance controller and RAIC are as follows.  Finding that a stiff controller performs better in rocky terrains than more compliant settings, we chose base impedance controller parameters sufficiently large to penetrate the slate material using the PDS trajectory.  We slowly increased K values until penetration was possible, which occurred at 750\,N/m for linear stiffness values and 80\,Nm/rad for rotational stiffness. Next we tuned damping values until the scoop's oscillations when hitting hard terrain were minimal. We found these damping values to be 330\,Ns/m for linear damping and 12\,Nms/rad.

Next, we set the parameters of the RAIC controller as a fixed function of the force and torque values that trigger protective stops in the UR5 arm. We did this by  firmly pushing on the tip of the end-effector until the arm triggered a protective stop and logging the magnitude of the force and torque values. We found these to be $F_\text{max}$ = 60 N and $T_\text{max}$ = 10 Nm respectively. We then set the cutoffs $c_i$ to be 25\% of their respective maxes and set the scaling values $s_i$ to be 1/(50\%) of their respective maxes. So, RAIC decouples the attractor point from the planned pose when 25\%-75\% of the max loads are encountered. For the feedback cutoffs, we set the cutoffs $c^{fb}_i$ to be 0 and set their scaling values $s^{fb}_i$ to be 1/(25\%) of their respective maxes. This is so that the attractor pose is pulled to the plan only when loads are low.

\subsection{Effects of Primitive Parameters}

\begin{figure}[tbp]
    \centering
    \includegraphics[width=\linewidth]{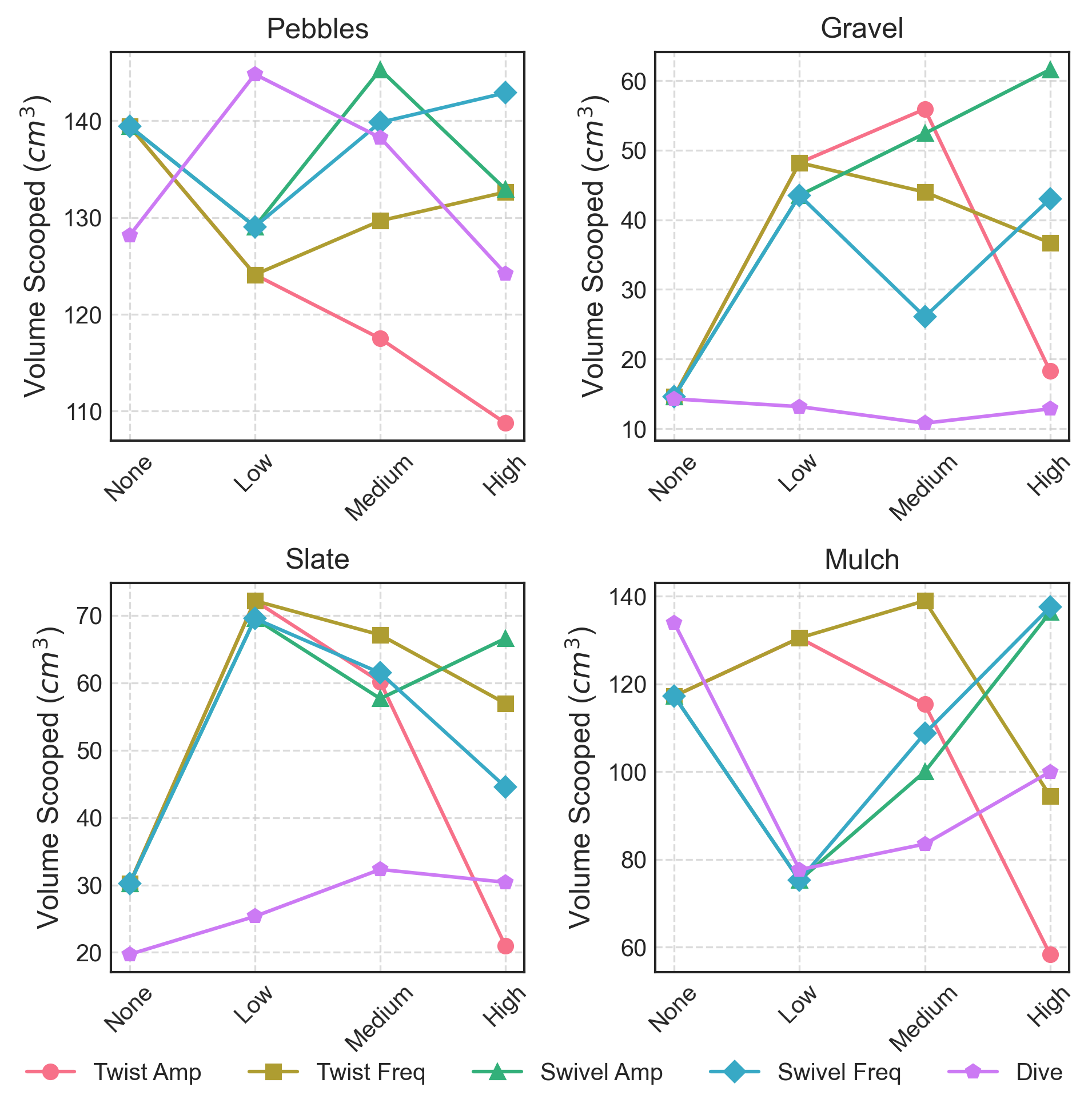}
    \caption{Parameter exploration showing how primitive parameter settings are related to excavated volume.  Dive sweep = None is equivalent to the standard penetrate-drag-scoop trajectory.}
    \label{fig:parameter_tuning}
\end{figure}
Next, we explore the effects of parameter settings on our proposed primitive trajectories. On each of our four terrains, we performed parameter sweeps along the five proposed parameters: swivel amplitude/frequency, the twist amplitude/frequency, and the dive curve. Each scoop maintains static values for depth of .07\,m, drag length of .25m, and attack angle of 5$\pi$/6 radians. 

For each setting we performed 6 (12 on slate for verification purposes) scoops and measured the excavated volume, counting P-stop events as 0. For each parameter, we swept through the primitive at zero, {\em Low}, {\em Medium}, and {\em High} setting. The swivel amplitude sweep considers $A_{\text{swivel}}$ at $\pi$/16, $\pi$/8, and 3$\pi$/16\,rad while keeping $f_\text{swivel}$ at the {\em Low} frequency ($\pi$\,rad/sec). The twist amplitude sweep consideres $A_{\text{twist}}$ at $\pi$/12, $\pi$/6 and $\pi$/4\,rad respectively while keeping $f_{\text{twist}}$ at the same {\em Low} frequency. Both swivel and twist frequency parameter sweeps vary $f_\cdot$ amongst $\pi$, 2$\pi$, and 4$\pi$\,rad/sec while maintaining {\em Low} amplitude. For the dive curve, $s$ is varied between 0.33, 0.66, and 1.0.  All other parameters are set to 0. Fig.~\ref{fig:parameter_tuning} shows the results from these sweeps.

We found that oscillatory movements during the loading cycle improved the average volume scooped in the difficult Slate and Gravel terrains, while showing little improvement in easier terrains such as Pebbles. For Gravel and Slate, we found that medium-high settings for swivel helped break up frictional forces that caused particles to jam. We also found that twisting at small amplitudes but low-medium frequencies helped improve volume scooped in materials with larger particle sizes. However, twisting at higher amplitudes in lost some material due the roll of the scoop becoming too aggressive, spilling material from the sides.

As for the dive curve, we found that although maintaining a pitch tangential to the trajectory improved performance when combined with a swivel or twist motion, the value of dive curve itself did not affect the results much. In Slate, we can see a minor improvement when we varied $s$ from 0 to 1, however in the other materials we saw an unsubstantial difference in volume scooped between the different values.

\subsection{Ablation studies}
We next perform ablation studies to examine the effects of our primitives and the RAIC controller (Tab.~\ref{tab:performance_comparison}). The four experimental conditions use were on the impedance controller using the PDS trajectory, impedance controller plus our scooping primitive trajectory, our RAIC controller using the PDS controller and our RAIC controller using our scooping primitive trajectory. For the scooping primitive trajectory, we used the best parameters described in our parameter tuning section. Both the impedance controller and RAIC controller use the tuned parameters found in our parameter tuning experiments of 750 N/m linear stiffness, 80 Nm/rad rotational stiffness, 320 Ns/m linear damping and 12 Nms/rad rotational damping. In each study we perform 12 (30 on slate for verification purposes) scoops per terrain in batches of 2 pre-defined penetration points. The scoop parameters used in these experiments are .07\,m depth, .25\,m drag length, and attack angle of 5$\pi$/6 radians.

We computed 3 metrics: volume scooped, P-stop rate, and completion percentage. Completion percentage is defined as the percentage of the trajectory executed before a P-stop occurred.  In the easy Pebbles terrain, standard methods work adeuqately well.  However, the RAIC controller helps prevent protective stops in harder terrains, especially Gravel and Slate. In these terrains, P-stops were frequently triggered, often during the penetration phase or early drag phase as indicated by the low completion percentages. Behavior in Mulch was also slightly improved using RAIC, although it is not clear whether these results are statistically significant due to the high variance in scooped volumes.

Beyond RAIC, our scooping primitives both helped reduce the number of P-stops as well as increased average scoop volume in most cases. In Gravel and Slate, oscillations broke the friction forces that caused the particles to jam as well as displaced particles that were interlocked with one another, increasing volume and reducing P-stop rate to nearly 0\%.  

\subsection{Hidden Obstacle Compliance}
Finally, we performed two qualitative experiments to show how the RAIC controller complies with hidden obstacles, as an excavator can often encounter tangled roots, large boulders, or other unseen obstructions in the terrain.

\begin{figure}[tbp]
\centering
\includegraphics[width=0.49\linewidth]{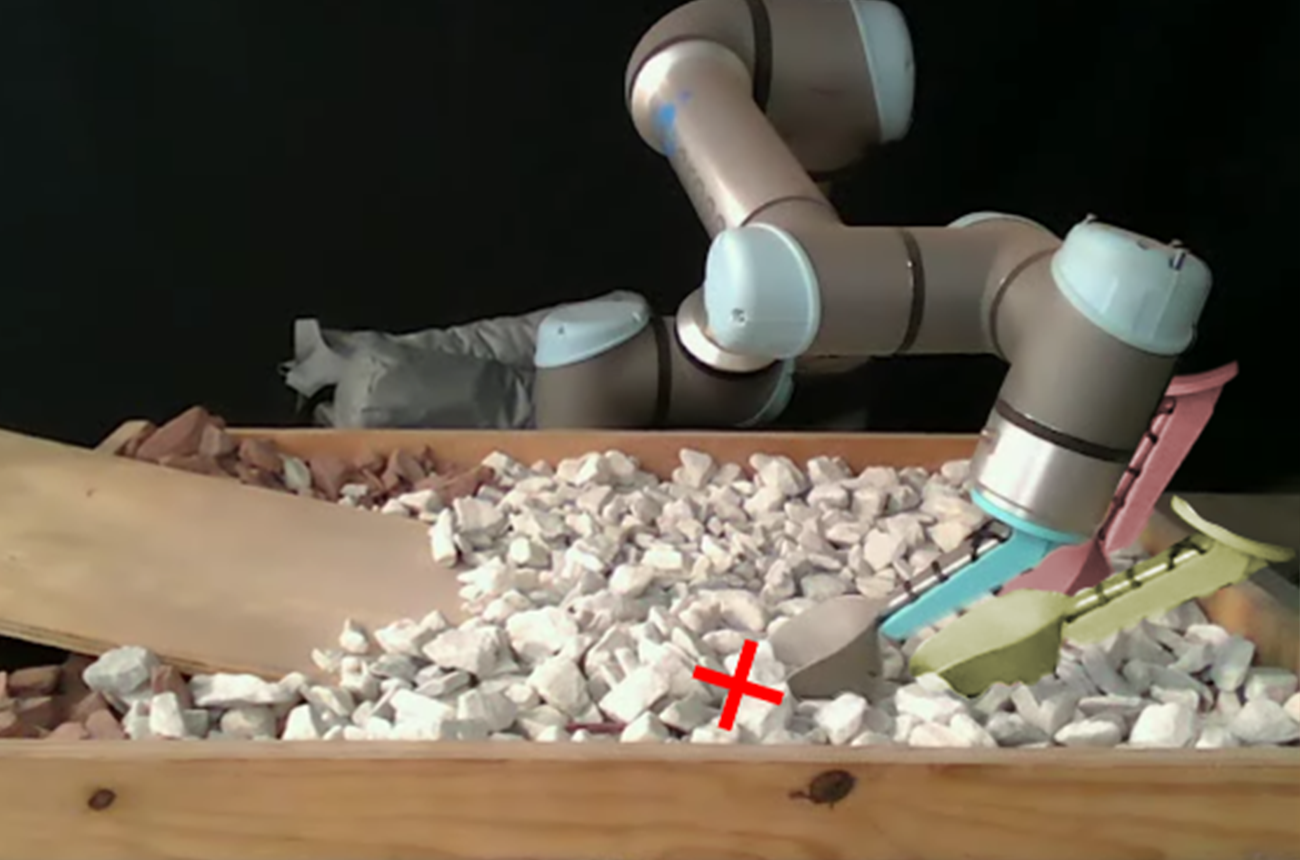} 
\includegraphics[width=0.49\linewidth]{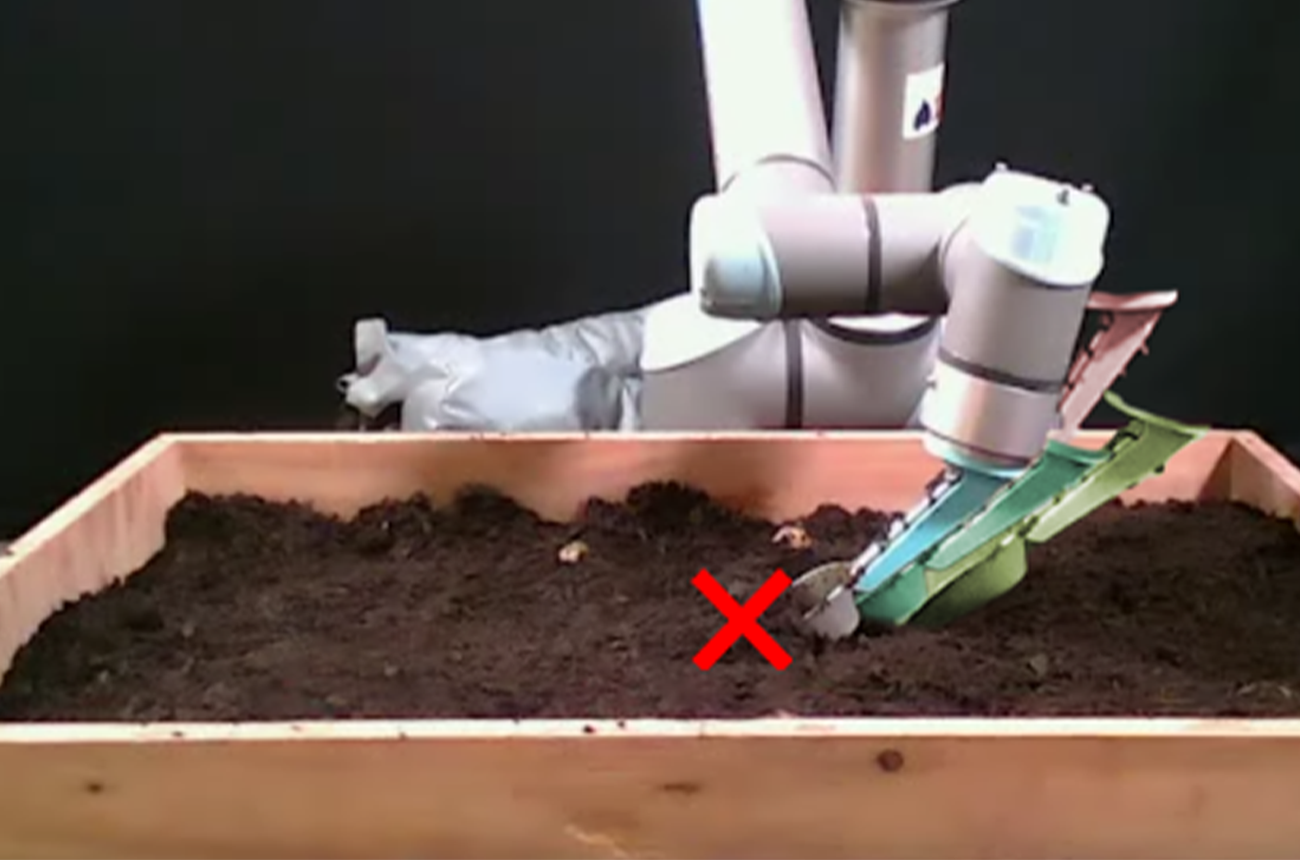}\\
\vspace{1mm}
\includegraphics[width=0.49\linewidth]{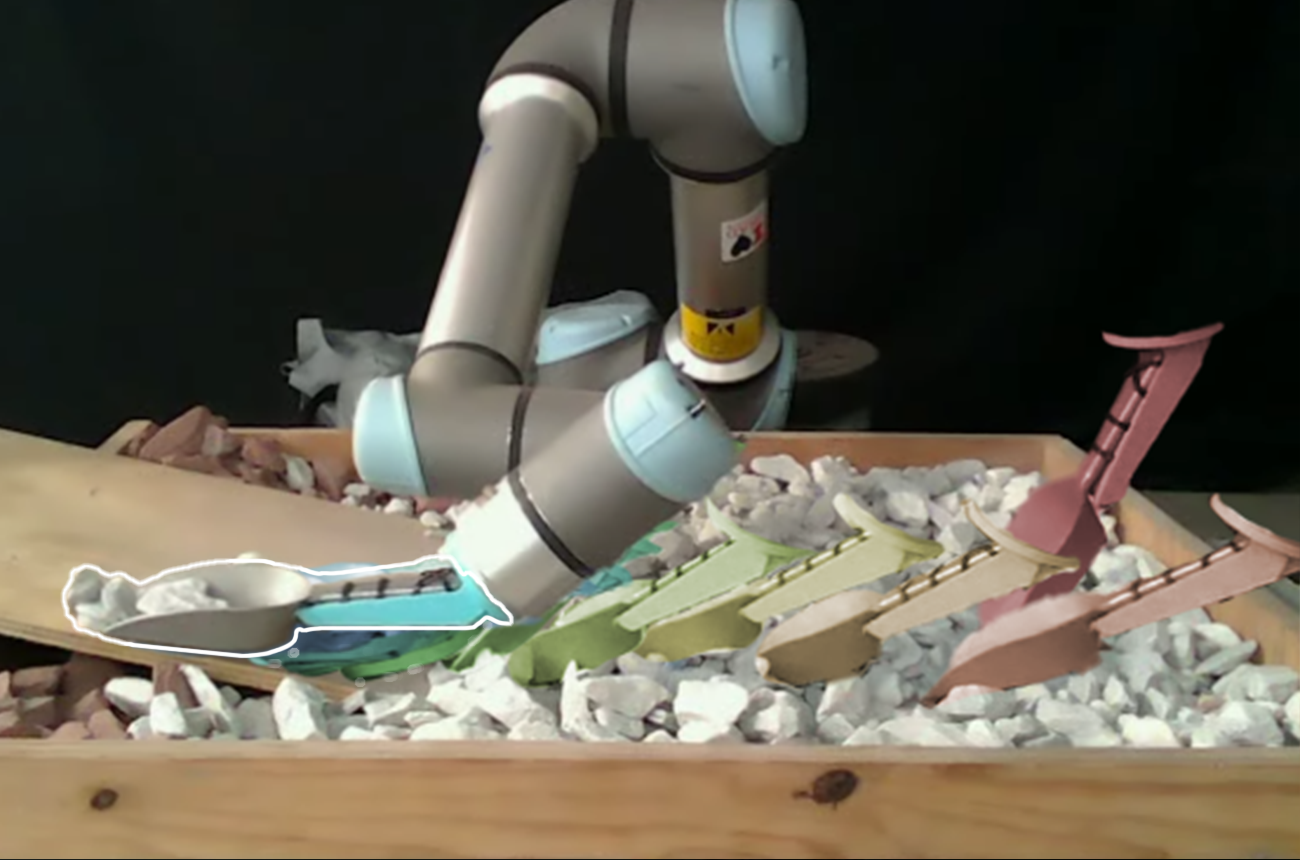} 
\includegraphics[width=0.49\linewidth]{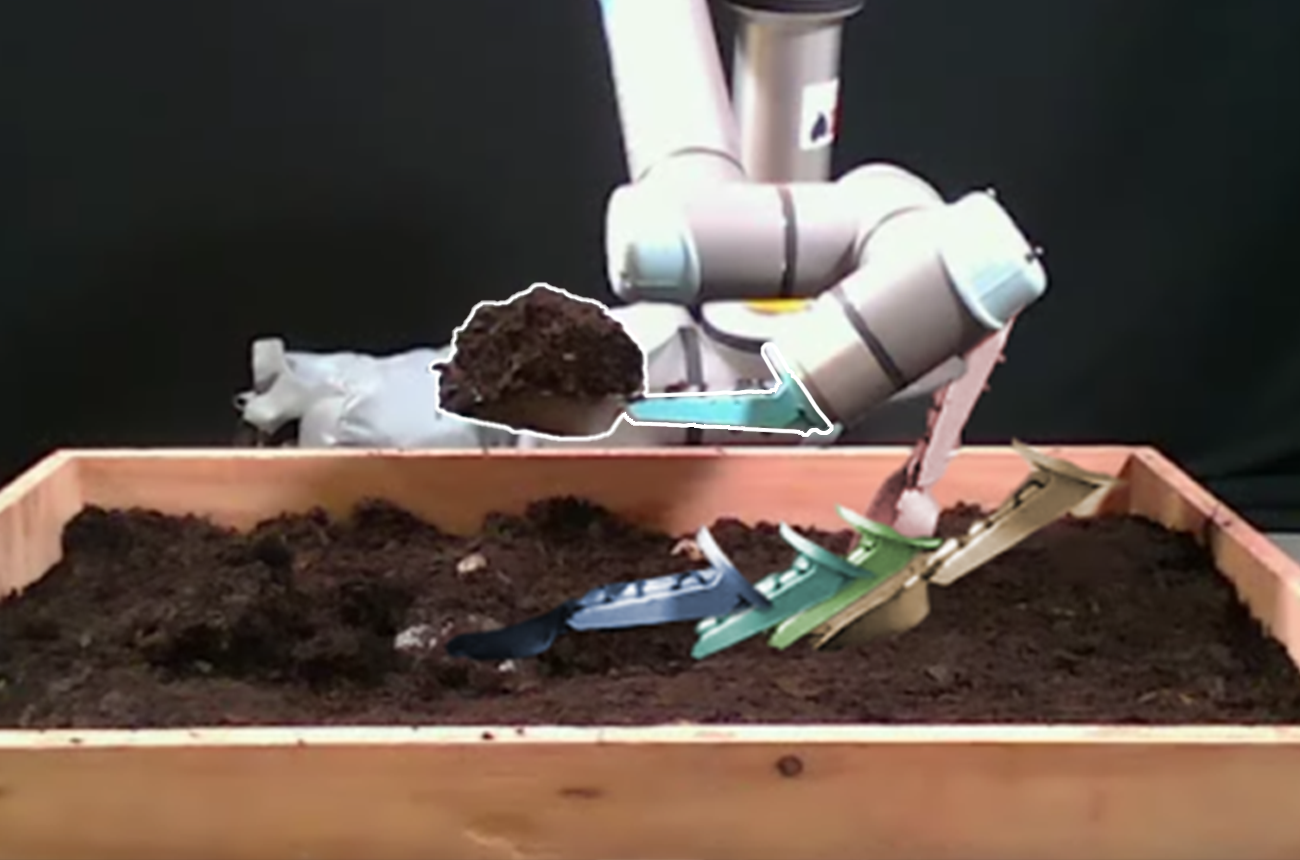} 
\caption{Testing on hidden obstacles, showing movement via strobe-effect. Left: a rigid slope is embedded in gravel. Right: a large rock embedded in soil. Top row: a standard impedance controller is unable to comply to these obstacles when jamming is encountered, triggering a protective stop. Bottom row: the RAIC controller successfully adapts to these obstacles.  }
\label{fig:hidden_obstacles}
\end{figure}

The first experiment embeds a fixed slope into gravel.
For this experiment, we set the depth to be .06\,m, drag length of .35\,m, and an attack angle of 5$\pi$/6 radians. Primitive settings include {\em High} $A_\text{swivel}$, {\em Low} $f_\text{swivel}$, and a dive curve $s=0.5$. Compared to a standard impedance controller, RAIC is able to comply to the slope and successfully excavate material (Fig.~\ref{fig:hidden_obstacles}, left).

The second experiment embeds a large stone into soft soil.  We created an indentation in the terrain, placed a large rock in the center of it, and refilled the soil to cover the rock, lightly compacting it.  The primitive parameters are unchanged from the ablation studies, using the dive primitive at dive curve $s=0.5$.  Compared to standard impedance control, which jams against the rock, RAIC is able to both slow down the trajectory and scoop underneath the rock, leading to the successful completion of the trajectory.(Fig.~\ref{fig:hidden_obstacles}, right)

\begin{table}[t]
    \centering
    \caption{Performance comparison of control methods across test terrains. Volume averages and standard deviations are presented. Bold value indicates best mean on a given terrain}
    \label{tab:performance_comparison}
    \small
    \begin{tabular}{@{}llccc@{}}
        \toprule
        Method & Material & \begin{tabular}[c]{@{}c@{}}Volume\\ (cm³)\end{tabular} & \begin{tabular}[c]{@{}c@{}}P-stop\\ rate (\%) \end{tabular} & \begin{tabular}[c]{@{}c@{}}Completion \\ \%\end{tabular} \\
        \midrule
        \multirow{4}{*}{\begin{tabular}[c]{@{}l@{}}Impedance +\\ PDS\end{tabular}} 
            & Pebbles & 134.5 $\pm$ 24.4 & 25 & 85 \\
            & Gravel  & 4.0 $\pm$ 4.7    & 100 & 21 \\
            & Slate   & 12.3 $\pm$ 3.2   & 100 & 27 \\
            & Mulch   & 119.0 $\pm$ 67.0 & 25 & 85 \\
        \midrule
        \multirow{4}{*}{\begin{tabular}[c]{@{}l@{}}Impedance +\\ Primitives\end{tabular}}
            & Pebbles & 142.0 $\pm$ 11.5  & {\bf 0} & {\bf 100} \\
            & Gravel  & 38.7 $\pm$ 44.3   & 66 & 51 \\
            & Slate   & 45.6 $\pm$ 59.8   & 66 & 52 \\
            & Mulch   & 106.4 $\pm$ 66.6  & 17 & 89 \\
        \midrule
        \multirow{4}{*}{\begin{tabular}[c]{@{}l@{}}RAIC + PDS\\ \end{tabular}}
            & Pebbles & 140.0 $\pm$ 15.3 & 8 & 95 \\
            & Gravel  & 18.7 $\pm$ 26.0  & 50 & 65 \\
            & Slate   & 17.6 $\pm$ 18.8  & 50 & 64 \\
            & Mulch   & {\bf 135.7 $\pm$ 59.9} & {\bf 0} & {\bf 100} \\
        \midrule
        \multirow{4}{*}{\begin{tabular}[c]{@{}l@{}}RAIC +\\ Primitives\end{tabular}}
            & Pebbles & {\bf 143.8 $\pm$ 9.2}  & {\bf 0} & {\bf 100} \\
            & Gravel  & {\bf 62.5 $\pm$ 31.9}  & {\bf 8} & {\bf 94} \\
            & Slate   & {\bf 68.6 $\pm$ 35.7}  & {\bf 7} & {\bf 95} \\
            & Mulch   & 97.0 $\pm$ 52.3  & 8 & 95 \\
        \bottomrule
    \end{tabular}
\end{table}

%%%%%%%%%%%%%%%%%%%%%%%%%%%%%%%%%%%%%%%%%%%%%%%%%%%%%%%%%%%%%%%%%%%%%%%%%%%%%%%%
\section{Conclusion}
This paper demonstrated the surprising effectiveness of oscillation and adaptive impedance control when excavating challenging terrains with large grain sizes and inter-grain entangling.  Oscillation is used to break up grains and avoid jamming, and adaptive impedance maintains measured progress and low forces even in cases of severe jamming.  We note that these techniques use capabilities that are commonly available in industrial robot arms but not in current excavators, whose arms are usually 4DOF devices that cannot roll or yaw about the bucket, and which require significant retrofitting to enable force feedback \cite{ETH}. Our work suggests the possibility of incorporating new approaches in the design of more effective excavation mechanisms (both  automated and manually controlled).  Although our experiments indicate that appropriate parameter settings can greatly improve the volume of excavated materials, we do not yet have an automated technique for identifying optimal parameters for a given terrain.  In future we plan to investigate using material identification to select appropriate parameters a priori, as well as using force and/or vision feedback to adapt oscillations on the fly.

%\addtolength{\textheight}{-12cm}   % This command serves to balance the column lengths
                                  % on the last page of the document manually. It shortens
                                  % the textheight of the last page by a suitable amount.
                                  % This command does not take effect until the next page
                                  % so it should come on the page before the last. Make
                                  % sure that you do not shorten the textheight too much.

%%%%%%%%%%%%%%%%%%%%%%%%%%%%%%%%%%%%%%%%%%%%%%%%%%%%%%%%%%%%%%%%%%%%%%%%%%%%%%%%

%%%%%%%%%%%%%%%%%%%%%%%%%%%%%%%%%%%%%%%%%%%%%%%%%%%%%%%%%%%%%%%%%%%%%%%%%%%%%%%%

%%%%%%%%%%%%%%%%%%%%%%%%%%%%%%%%%%%%%%%%%%%%%%%%%%%%%%%%%%%%%%%%%%%%%%%%%%%%%%%%

\section*{ACKNOWLEDGMENT}
This work is partially supported by NASA Grant \#80NSSC21K1030 and NSF Grant \#FRR-2409661.

\bibliographystyle{IEEEtran}
\bibliography{references}

\end{document}